\documentclass[conference]{IEEEtran}
\usepackage{graphicx}

\usepackage{booktabs}

\ifCLASSINFOpdf
\else
\fi

\usepackage{amsmath}
\usepackage{xurl}
\usepackage{multirow}

\begin{document}
%
\title{Toward Short-Term Glucose Prediction Solely Based on CGM Time Series
}

\author{\IEEEauthorblockN{Ming Cheng$^{*}$, Xingjian Diao$^{*}$, Ziyi Zhou$^{*\dag}$, Yanjun Cui, Wenjun Liu, Shitong Cheng}
\IEEEauthorblockA{$^1$Department of Computer Science, Dartmouth College, Hanover, NH 03755, USA\\
Email: ziyi.zhou.gr@dartmouth.edu}
}

\maketitle

\begin{abstract}
The global diabetes epidemic highlights the importance of maintaining good glycemic control. Glucose prediction is a fundamental aspect of diabetes management, facilitating real-time decision-making. Recent research has introduced models focusing on long-term glucose trend prediction, which are unsuitable for real-time decision-making and result in delayed responses. Conversely, models designed to respond to immediate glucose level changes cannot analyze glucose variability comprehensively. Moreover, contemporary research generally integrates various physiological parameters (e.g. insulin doses, food intake, etc.), which inevitably raises data privacy concerns. To bridge such a research gap, we propose TimeGlu -- an end-to-end pipeline for short-term glucose prediction solely based on CGM time series data. We implement four baseline methods to conduct a comprehensive comparative analysis of the model's performance. Through extensive experiments on two contrasting datasets (CGM Glucose and Colás dataset), TimeGlu achieves state-of-the-art performance without the need for additional personal data from patients, providing effective guidance for real-world diabetic glucose management.

\end{abstract}
\newcommand\blfootnote[1]{%
  \begingroup
  \renewcommand\thefootnote{}\footnote{#1}%
  \addtocounter{footnote}{-1}%
  \endgroup
}

\blfootnote{$^*$Equal contribution. $^\dag$Corresponding author.}

\IEEEpeerreviewmaketitle

\section{Introduction}

Approximately 537 million people globally are currently living with diabetes~\cite{9871646}, and this number is expected to continue increasing~\cite{saeedi2019global}. This escalating trend highlights the complexities of managing the condition, particularly the challenge of maintaining good glycemic control~\cite{bartolome2022computational} amid diverse factors~\cite{ahmad2014factors}. Consequently, there is an increased demand for accurate blood glucose prediction, a key aspect of diabetes management~\cite{vettoretti2020advanced}.

Blood glucose prediction, with the goal of forecasting future blood glucose levels to enable real-time decision-making in diabetes management, has gained substantial importance amid the growing diabetes epidemic~\cite{9871646,saeedi2019global} and evolving digital health technologies~\cite{solomon2020digital, bartolome2021glucomine} for effective glycemic control. This pursuit has notably advanced in prominence and relevance within the healthcare and medical research communities, influencing applications including anomaly detection \cite{wang2020guardhealth} and monitoring \cite{li2021fully, wang2024differential}, preventive healthcare~\cite{ng2022predicting}, wearable health technology~\cite{10124956, li2022smart, AlikhaniKoshkak2024SEAL}, and artificial pancreas development~\cite{nwokolo2023artificial}. 

Recent blood glucose prediction models in diabetes management typically fall into \textbf{\textit{two}} types.
The \textbf{\textit{first}} emphasizes the prediction of long-term blood glucose trends.  These models~\cite{liu2019long, jaloli2023long} provide an extensive view of glycemic control over time, crucial for strategic treatment and lifestyle adjustments~\cite{choe2022effects}. They are valuable in identifying long-term patterns~\cite{keum2018long} and potential complications~\cite{mezil2021complication} arising from sustained hyperglycemia or hypoglycemia. However, these models are less effective for real-time decision-making as they focus on long-term data analysis rather than current fluctuations. In addition, the emphasis on long-term trends can lead to delayed responses to acute changes in glucose levels, potentially missing immediate risks~\cite{jafar2016effect,mantovani2016severe}. The \textbf{\textit{second}} path particularly focuses on responding to immediate glucose level changes. Although these models~\cite{dassau2010real, ben2015identification} are effective for immediate adjustments, they tend to be reactive, concentrating on current glucose values and often lacking the capability to predict future trends or analyze patterns in glucose variability. 

Moreover, many current models in glucose level prediction often integrate various physiological parameters~\cite{bartolome2022computational}, such as insulin doses, food intake, and physical activity, alongside CGM data. They come with a significant drawback while showing efficacy: The reliance on multiple data sources can complicate their use in real-world settings due to data availability~\cite{avram2020digital} and patient compliance issues~\cite{cheng2023saic}. Additionally, such models cannot fully exploit temporal patterns in CGM data.

A notable gap in current research is the utilization of historical CGM data for forecasting short-term glucose trends. Predicting these trends solely based on CGM data is important for decision-making and proactive diabetes management~\cite{george2019proactive, zhou2024crossgp}.

Addressing these challenges is crucial not only for individual healthcare management~\cite{pasquel2021management} but also for broader medical research~\cite{zhou2021doseguide,zhang2023doseformer}, which aims to develop more efficient monitoring and intervention methods. Consequently, one crucial question has arisen: Is it feasible to accurately forecast high-frequency blood glucose level trends over short periods using only CGM data?

To overcome the challenges above, we propose \textbf{TimeGlu} -- a one-stage end-to-end pipeline for short-term glucose prediction solely based on CGM time series.

In summary, our contribution is threefold:
\begin{enumerate}
    \item {
    \textbf{One-stage end-to-end pipeline.} 
    We propose TimeGlu -- an effective one-stage end-to-end pipeline that forecasts the high-frequency blood glucose level trend over a short time series manner.
    }
    \item {
    \textbf{Effective high-frequency glucose level prediction.} 
    TimeGlu enables forecasting the high-frequency blood glucose level trend over a short time series manner.
    }
    \item {
    \textbf{State-of-the-art performance.} 
    We conduct extensive experiments on the CGM Glucose dataset\footnote{\url{https://github.com/DigitalBiomarkerDiscoveryPipeline/Case_Studies}} and the Colás dataset \cite{colas2019detrended} with various methods to do time series-based glucose prediction. To the best of our knowledge, ours is the first study to do glucose prediction solely based on time series on the Colás dataset. Our model can consistently accurately predict high-frequency blood glucose level trends across different conditions.
    }
\end{enumerate}

\section{Methods}
To obtain accurate glucose prediction results, we employ four different methods for time series-based glucose prediction. Afterward, we analyze and compare the results of these methods to determine the optimal glucose prediction model.

\subsection{Exponential Smoothing}
The Exponential Smoothing method \cite{gardner1985exponential} is widely used in accurate prediction for short-term time series, 
making it suitable for glucose prediction. 
It assigns greater importance to recent observations, with exponentially diminishing weights for observations that are farther in the past. 
Considering that short-term glucose changes are trending over time, we adopt the Double Exponential Smoothing (DFS) model \cite{heckert2002handbook} for glucose prediction. 
Formally, given $X = \{x_1, x_2, ..., x_n\}$ as the observed time series, for time $t > 0$, its smoothed value ($s_t$) and the best estimate of the trend ($b_t$) are expressed as follows:
\begin{align}
    s_t &= \alpha x_t + (1 - \alpha)(s_{t-1} + b_{t-1}) \\
    b_t &= \beta (s_t - s_{t-1}) + (1-\beta)b_{t-1}
\end{align}

where $\alpha$ and $\beta$ are time/trend smoothing factor, and initially $s_0 = x_0$, $b_0 = x_1 - x_0$. 

Therefore, the prediction value ($\Tilde{x}_{t+k}$) at time ($t+k$) is expressed as: 
\begin{equation}
    \Tilde{x}_{t+k} = s_t + kb_t
\end{equation}

\subsection{Auto ARIMA}
As a commonly employed method for time series data forecasting, ARIMA is characterized by three order parameters: $p, d, q$. Generally, for an \textit{ARIMA(p, d, q)} model, $p$ indicates the order of the autoregressive part, $d$ stands for the degree of first differencing involved, and $q$ refers to the order of the moving average part \cite{awan2020prediction}. Formally, given $X = \{x_1, x_2, ..., x_n\}$ as the observed time series, the predicted value ($\Tilde{x_t}$) at time $t$ by the model \textit{ARIMA(p, q)} can be expressed as \cite{hyndman2018forecasting}: 
\begin{equation}
\label{arimaeq}
    \Tilde{x_t} = \delta + \sum_{i=1}^{p}\phi_ix_{t-i} + \sum_{j=1}^{q}\theta_i\epsilon_{t-j} + \epsilon_t
\end{equation}
where $\epsilon_t$ is the error term, and $\phi_i, \theta_i, \delta$ are parameters.

Instead of choosing the parameters in Equation \ref{arimaeq} manually, modern research mainly utilizes Auto ARIMA \cite{awan2020prediction, melard2000automatic, tran2004automatic} to find the best parameter combinations automatically by comparing the AIC/BIC criterion \cite{vrieze2012model} of different models, which significantly reduces the required human resources for selecting the parameters of the ARIMA model.

\subsection{BATS and TBATS}
Based on the previously mentioned Exponential Smoothing method and the ARMA model, the BATS model \cite{de2011forecasting} considering the Box-Cox Transformation \cite{sakia1992box} for time series forecasting is proposed. Specifically, befitting from the ability to deal with non-linear data and de-correlate the time series data, BATS showcases the improvement in time series prediction performance \cite{de2010automatic} over the methods mentioned above. 

Considering the limitations of BATS when the data seasonality is complex and high frequency (e.g. seasonalities of time in a day), BATS with Trigonometric Seasonality (TBATS) \cite{de2011forecasting} is constructed. Experimental results (Section \ref{sec:exp}) on the CGM dataset demonstrate the advantages of the TBATS over BATS in dealing with time series data.

\begin{figure*}[tb]
  \centering
  \includegraphics[width=0.9\textwidth]{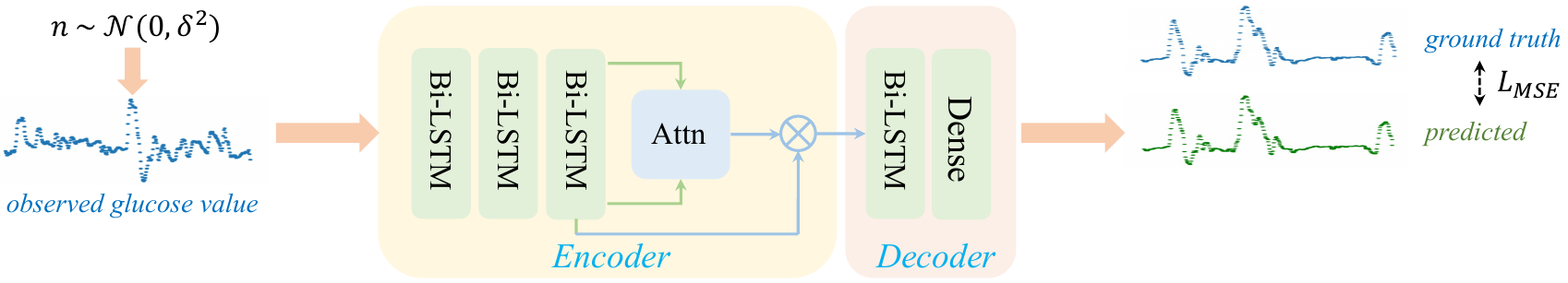}
  \caption{\textbf{
  Overview of the proposed TimeGlu pipeline.} 
  Noise with standard normal distribution is applied to the input data for data augmentation, which is then input into the encoder-decoder-based architecture. To precisely and comprehensively extract the time series features, a sequential Bi-LSTM structure with an Additive Attention \cite{bahdanau2014neural} module is designed. The high-dimensional features will be input into a lightweight decoder to generate the predicted glucose value. 
    An MSE loss is constructed for pipeline training. 
  }
  \label{fig:pipeline}
\end{figure*}

\subsection{Neural Networks -- TimeGlu}
Considering the powerful deep feature extraction capability of neural network models and inspired by the mainstream encoder-decoder designs \cite{lyu2020lstm, hu2020dasgil, zhang2021novel}, we design a novel encoder-decoder-based pipeline for glucose prediction solely based on CGM time series data, as shown in Figure \ref{fig:pipeline}.

Assume $x$ as the observed time series glucose data, the noise with normal distribution $n \sim (0, \delta^2)$ is applied for data augmentation, considering the variability of glucose changes across the temporal domain. The augmented data is then input into TimeGlu to obtain the predicted data $\Tilde{x}$, as expressed below:
\begin{equation}
    \Tilde{x} = D({E}(x + n)), \quad n \sim \mathcal{N}(0, \delta^2)
\end{equation}
where $E$ and $D$ indicate the encoder and decoder of TimeGlu, respectively. 

To train the pipeline, the Mean Square Error (MSE) \cite{allen1971mean} is constructed between the predicted glucose value and the ground truth:
\begin{equation}
    L_{MSE} = {||\Tilde{x} - y||}^2
\end{equation}
where $y$ indicates the ground truth. 

\subsubsection{Encoder Architecture}
The encoder module is designed with sequential Bidirectional LSTM (Bi-LSTM) \cite{siami2019performance} blocks followed by an Additive Attention \cite{bahdanau2014neural} layer. 

Specifically, Bi-LSTM processes sequences bidirectionally to learn past and future contexts simultaneously, with each direction containing an LSTM network. 
Let $X = \{x_1, x_2, ..., x_t\}$ denote the input sequence to the Bi-LSTM, where each $x_t$ is the data point at time step $t$. This sequence can consist of various forms of time-sensitive data including physiological measurements in healthcare applications. Formally, the hidden states of the forward LSTM ($h_t^f$) and backward LSTM ($h_t^b$) at timestamp $t$ are expressed as follows:
\begin{equation}
        h_t^f = LSTM(x_t, h_{t-1}^f), \quad h_t^b = LSTM(x_t, h_{t+1}^b)
\end{equation}

The output of Bi-LSTM at timestamp $t$ is then combined through the concatenation operation:
\begin{equation}
    h_t = h_t^f \oplus h_t^b
\end{equation}

Inspired by existing attention-based structures \cite{zhou2024crossgp, wu2021fastformer, diao2023av}, an Additive Attention module \cite{bahdanau2014neural} is applied to instruct the model to selectively focus on the relevant parts of the input sequence, following the Bi-LSTM output. Since the attention mechanism is employed within $h_t$ (as shown in Figure \ref{fig:pipeline}), this module dynamically lets the model capture key components of the features.  Afterward, inspired by the fusion operation introduced in \cite{diao2023ft2tf}, the output of the attention is then concatenated together with the Bi-LSTM output to assign adaptive weights to different parts of the learned features. Therefore, the output of the encoder ($x'_t$) at timestamp $t$ can be expressed as: 
\begin{equation}
    x'_t = Con(h_t, attn(h_t, h_t))
\end{equation}
where $Con$ represents the concatenation operation.

\subsubsection{Decoder Architecture}

A lightweight decoder containing Bi-LSTM and a Dense layer is constructed to generate the predicted value. 
Benefiting from the deep and dense blocks of the encoder, TimeGlu can accurately predict glucose values with only a lightweight decoder.
In contrast to some complex decoder modules, this design allows TimeGlu to infer in real time, enabling its generalization to practical applications.

\begin{figure*}[tb]
  \centering
  \includegraphics[width=\textwidth]{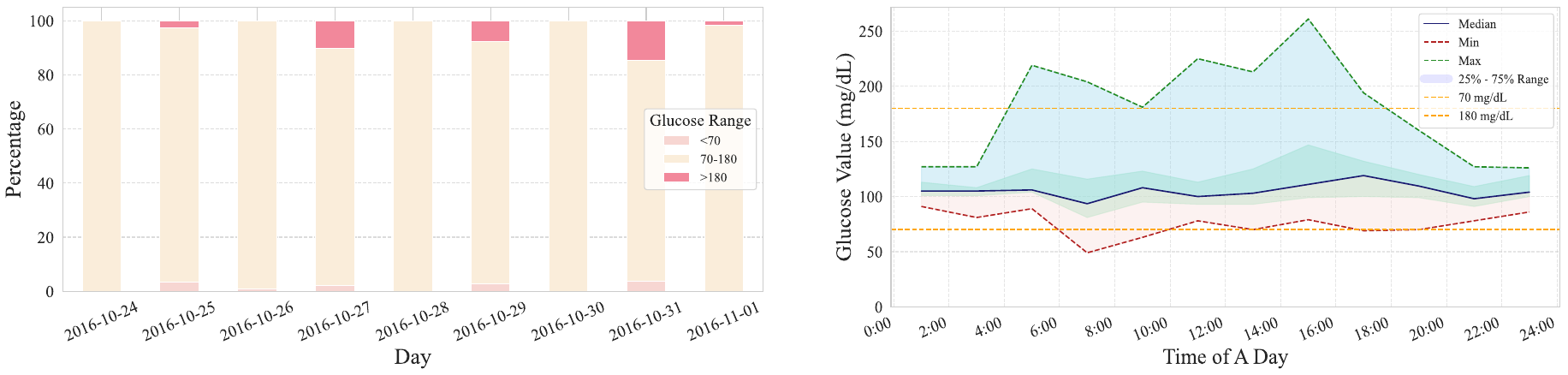}
  \caption{\textbf{
   Visualization of the CGM Glucose dataset.} 
   \textbf{Left: } Three different ranges are demonstrated across days within one week.
   \textbf{Right: } TIR/TAR/TBR visualization of the dataset. Multiple levels (median, min, max, etc.) of glucose values are illustrated.
  }
  \label{fig:data1}
\end{figure*}

\begin{figure*}[tb]
  \centering
  \includegraphics[width=\textwidth]{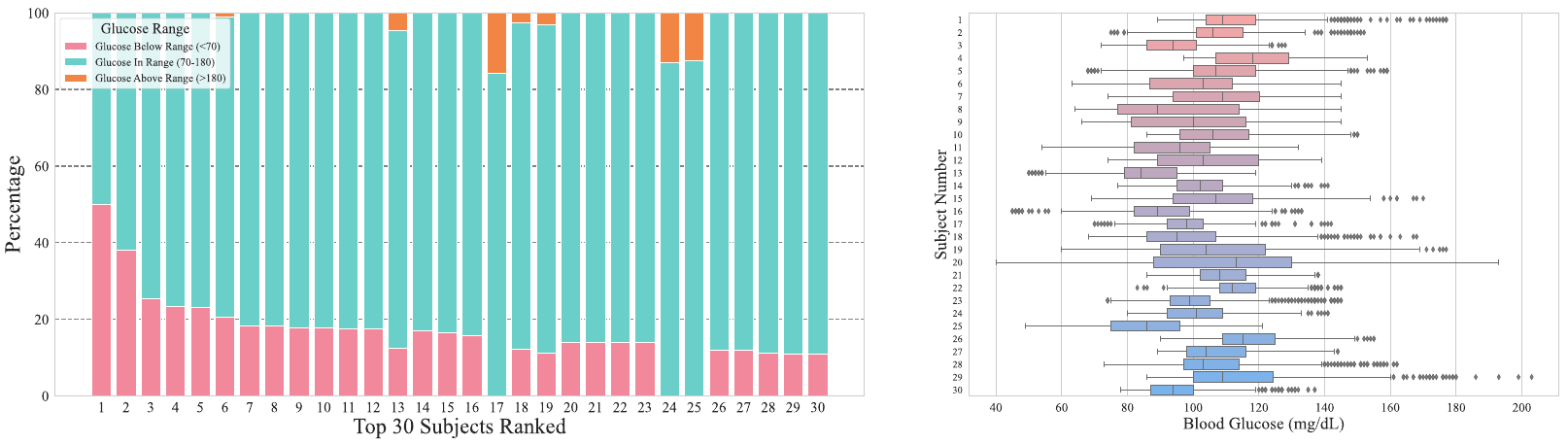}
  \caption{\textbf{
   Visualization of the Colás dataset.} 
   \textbf{Left: } 30 subjects with the most abnormal blood glucose situations are shown as an example.
   \textbf{Right: } 30 randomly selected subjects are illustrated through the box plot to showcase the variety and data diversity of the dataset.
  }
  \label{fig:data2}
\end{figure*}

\begin{table}[]
\centering
\caption{Summary of CGM Glucose Dataset and Colás Dataset}
\label{summary}
\begin{tabular}{@{}c|c|c|c@{}}
\toprule
\textbf{Dataset}             & \textbf{Attribute}    & \textbf{Mean $\pm$ STD} & \textbf{Range}          \\ \midrule
\multirow{3}{*}{\begin{tabular}[c]{@{}c@{}}CGM\\Glucose\end{tabular}} & Number of Data Points & 2147                    & -                       \\
                             & Time Range (day)            & 9                       & - \\
                             & Glucose Value (mg/dL)         &       111.9 $\pm$ 28.8                  &      49.0 - 261.0 
                     \\ \midrule
\multirow{6}{*}{Colás}       & Subjects / Data Points    & 208 / 114,912                     & -                       \\
                            & Time Range (year)    & -                     & 2012-2015                       \\
                             & Gender Ratio (F/M)    & 105 / 103               & -                       \\
                             & Age (yrs)                  & 59.6 $\pm$ 10.1         & 29 - 88                 \\
                             & BMI                   & 30.0 $\pm$ 4.7          & 18.1 - 48.7             \\
                             & Follow-up Range       & 979.8 $\pm$ 371.9       & 176.0 - 2211.0          \\ \bottomrule
\end{tabular}
\end{table}

\section{Experiments}
\label{sec:exp}

\subsection{Dataset Description}
We conduct extensive experiments on two publicly available datasets: the CGM Glucose dataset and the Colás dataset. 
The CGM Glucose dataset is collected from a single person's week-long CGM data which has been de-identified and time-shifted for privacy protection, while the Colás dataset includes 208 patients from the outpatient clinic of hypertension and vascular risk of the University Hospital of Móstoles in Madrid from January 2012 to May 2015 \cite{colas2019detrended}. 

\begin{figure*}[h]
  \centering
  \includegraphics[width=\textwidth]{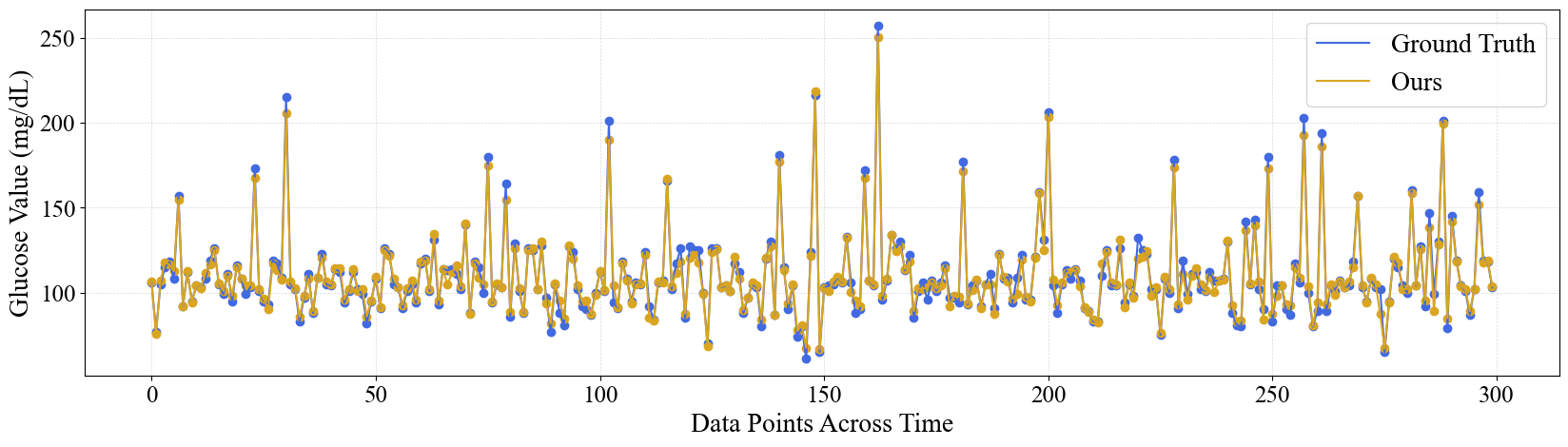}
  \caption{\textbf{
  Visualization of glucose value prediction of TimeGlu on the CGM Glucose dataset.} 
  TimeGlu is capable of accurately predicting glucose trends and specific values over time on the CGM Glucose dataset.
  }
  \label{fig:results1}
\end{figure*}

\begin{figure*}[h]
  \centering
  \includegraphics[width=\textwidth]{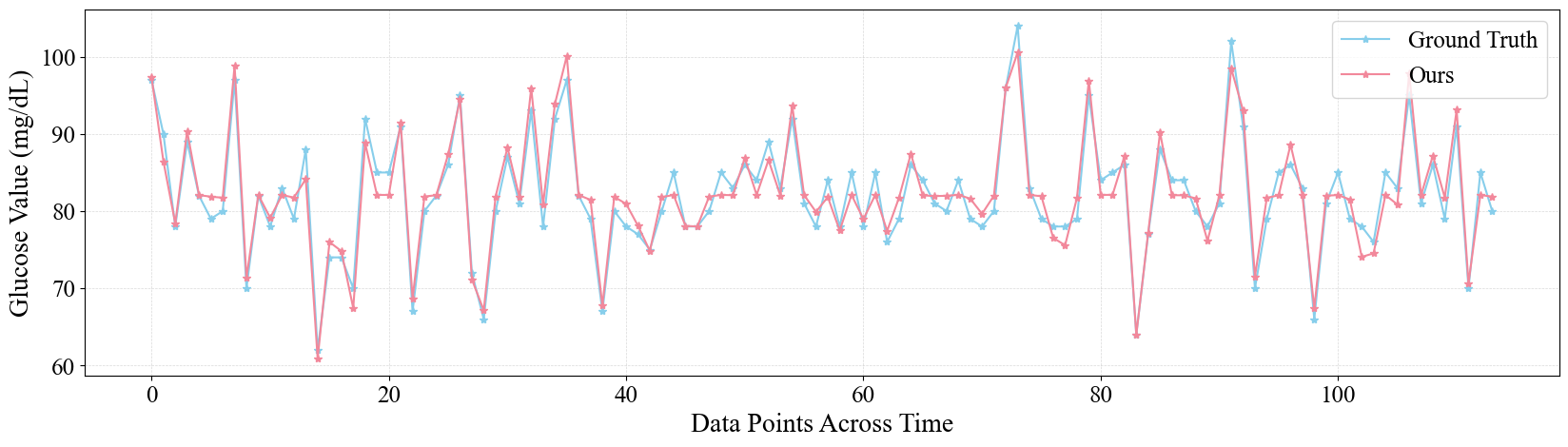}
  \caption{\textbf{
  Visualization of glucose value prediction of TimeGlu on the Colás dataset.} 
  For more clear demonstration, 114 data points are illustrated in the figure. 
  For large-scale datasets, TimeGlu provides robust generalization capabilities to accurately predict blood glucose trends as well as glucose values. 
  }
  \label{fig:results2}
\end{figure*}

\subsection{Dataset Analysis}
\subsubsection{CGM Glucose Dataset}
The dataset is summarized in Table \ref{summary}.
This week-long dataset contains $2147$
data points of a single person, ranging from October 24th 
to November 1st in 2016.
The glucose value of the selected person is relatively diverse, ranging from $49.0$
to $261.0$
including normal/abnormal states. 
The left part of Figure \ref{fig:data1}
illustrates glucose changes more intuitively throughout the day, where the glucose value is beyond the normal range in the daytime (from 4:00 AM to 6:00 PM) and 
reaches the highest at around 5:00 AM and 3:00 PM. 
The glucose changes throughout the entire week are shown in the right part of Figure \ref{fig:data1}, 
where a markedly abnormal range in glucose can be observed on October 27th and 31st.

\subsubsection{Colás Dataset}
The Colás dataset, as a relatively large-scale one exhibiting diversity, involves $114,912$ 
datapoints from $208$ subjects in total, as summarized in Table \ref{summary}.
Moreover, it is collected from 103 male and 105 female patients aged from $29$ to $88$, 
making it a challenging dataset for glucose prediction. 
Figure \ref{fig:data2} demonstrates an example of glucose range from 30 randomly selected patients, where the variety of data points is evidently to be proved. 
This gender-balanced dataset saturated with data diversity makes it an ideal dataset for evaluating the model.

\subsection{Evaluation Metrics}
To evaluate the performance of each method, we employ two objective evaluation metrics: Mean Absolute Error (MAE) \cite{hyndman2006another} and Mean Absolute Percentage Error (MAPE) 
\cite{khair2017forecasting}, to assess the accuracy of model predictions. Suppose $y$ and $\Tilde{y}$ are actual and predicted data points, respectively, the MAE and MAPE criterion can be expressed as below:
\begin{equation}
MAE = \frac{\sum {|y - \Tilde{y}|}}{n}, \quad 
    MAPE = \frac{\sum \frac{|y - \Tilde{y}|}{y}}{n}
\end{equation}
where $n$ indicates the total number of data points.
MAE treats overestimation and underestimation equally and provides a balanced view of how well a model is performing in terms of both positive and negative errors. In addition, MAPE quantifies the average magnitude of errors produced by the model, working as the most common metric of model prediction accuracy. 

\subsection{Quantitative Results}
The quantitative results on the CGM Glucose dataset and Colás dataset are shown in Table \ref{table:result}.

As indicated in the table, TimeGlu showcases a significant advancement in predicting glucose levels compared to the existing models (DFS, BATS, TBATS, and Auto ARIMA) across two datasets. In the comparative analysis, it is evident that TimeGlu not only outperforms these models but also sets a new standard in glucose level prediction.

\begin{table}[]
\centering
\caption{Comparison of Glucose Value Predictions By Multiple Methods on Different Datasets}
\label{table:result}
\begin{tabular}{@{}c|cc|cc@{}}
\toprule
\multirow{2}{*}{\textbf{Methods}} & \multicolumn{2}{c|}{\textbf{CGM Glucose}} & \multicolumn{2}{c}{\textbf{Colás}} \\
                                  & \textbf{MAE $\downarrow$}        & \textbf{MAPE $\downarrow$}       & \textbf{MAE $\downarrow$}     & \textbf{MAPE $\downarrow$}   \\ \midrule
DFS  \cite{heckert2002handbook}                             & 413.12              & 3.699               & 388.84           & 2.868           \\
BATS \cite{de2011forecasting}                             & 32.58               & 0.288               & 50.60            & 0.384           \\
TBATS \cite{de2011forecasting}                            & 29.22               & 0.247               & 21.36            & 0.156           \\
Auto ARIMA  \cite{awan2020prediction}                      & 25.62               & 0.199               & 20.99            & 0.154           \\ \midrule
\textbf{TimeGlu (Ours)}              & \textbf{2.99}       & \textbf{0.027}      & \textbf{13.49}   & \textbf{0.118}  \\ \bottomrule
\end{tabular}
\end{table}

\subsubsection{Accuracy}
In terms of accuracy, as measured by the MAE, TimeGlu exhibits remarkable performance. On the CGM Glucose dataset, TimeGlu achieves an MAE of 2.99, showing an improvement of over 99.28\% compared to DFS (the least MAE of 413.12), 90.83\% compared to BATS (32.58), 89.77\% compared to TBATS (29.22), and 88.33\% compared to Auto ARIMA (25.62). Similarly, for the Colás dataset, TimeGlu's MAE of 13.49 represents improvements ranging from 35.75\% to 96.53\% compared to other models. 

These results indicate that TimeGlu is remarkably more accurate in its predictions. The significant reduction in MAE suggests that TimeGlu is capable of closely tracking the intricate patterns of glucose levels, which is crucial for reliable diabetes management and treatment planning. The exceptional performance of our model can be largely attributed to the Bi-LSTM structure it employs. Bi-LSTM processes data in both forward and backward directions, allowing the model to learn from complex temporal dynamics of both past and future.

\subsubsection{Adaptability}
Regarding adaptability, assessed through MAPE, TimeGlu achieves 0.027/0.118 on the CGM Glucose and the Colás dataset, reflecting improvements from 86.43\% to 99.27\% and 23.38\% to 95.89\% over other models, respectively. 
As a relative error metric, MAPE can reflect the model's ability to adapt to different scales and conditions of blood glucose levels. This adaptability is essential for personalized treatment based on individual blood glucose patterns.

\subsection{Qualitative Results}

The qualitative results of our prediction are shown in Figure \ref{fig:results1} and \ref{fig:results2}, where it is evidently to prove the model generalization and robustness across datasets. 
Specifically, for a relatively small dataset where week-long data points are collected, TimeGlu generates accurate predictions since the golden line and blue line consistently overlap over time. Particularly, TimeGlu offers precise predictions for both normal values (e.g. $\sim 100$ mg/dL) and outliers (e.g. $\sim 250$ mg/dL), showing the robustness of our model in various situations. 

For a large-scale (Colás) dataset consisting of 208 subjects with diverse genders, ages, and BMIs, TimeGlu produces remarkable prediction results on trends and values, where the pink line and blue line are closely aligned.
Although when blood glucose values are outside of the normal range ($\sim 60$ mg/dL), TimeGlu remains highly predictive at precise time stamps. 
This demonstrates the potential of our model to perform glucose prediction for timely intervention in real-life applications and provide guidance for the physical health of human beings.

\subsection{Ablation Studies}

\begin{table}[]
\centering
\caption{Ablation Studies on The Structure of TimeGlu Pipeline}
\label{table:ablation}
\begin{tabular}{@{}c|cc@{}}
\toprule
\textbf{Methods}               & \textbf{MAE $\downarrow$}  & \textbf{MAPE $\downarrow$}  \\ \midrule
LSTM$_{Encoder}$                   & 4.51          & 0.041          \\
LSTM$_{Decoder}$                & 3.52          & 0.032          \\
w/o Attention                  & 3.40          & 0.030          \\ 
\textbf{TimeGlu (Ours)} & \textbf{2.99} & \textbf{0.027} \\ \bottomrule
\end{tabular}
\end{table}

\begin{figure}[tb]
  \centering
  \includegraphics[width=\linewidth]{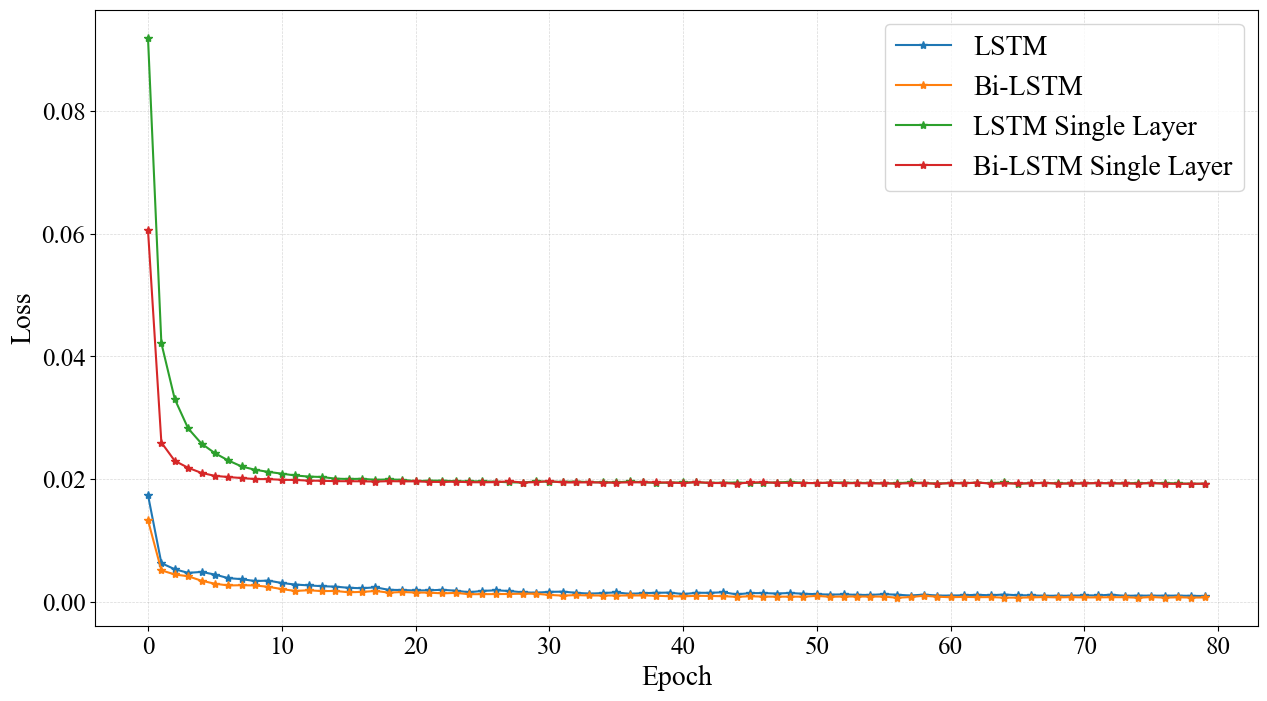}
  \caption{\textbf{
  Ablation studies on different encoder backbones. } 
  The training losses as epoch changes with different encoder backbones are shown.
  }
  \label{fig:ablation_loss}
\end{figure}

The ablation studies on pipeline design are shown in Table \ref{table:ablation} and Figure \ref{fig:ablation_loss}, where the effectiveness of the model structure is proved from multiple perspectives. 
In Table \ref{table:ablation}, LSTM$_{Encoder}$ and LSTM$_{Decoder}$ indicate that we replace Bi-LSTM with LSTM in encoder/decoder, where a significant decrease of accuracy can be observed (e.g. MAE error increases from $2.99$ to $4.51$ with LSTM in the encoder). 
Since MAE and MAPE criteria measure accuracy and interpretability, utilizing features from both directions (forward and backward) contributes to learning comprehensive representations from the time series data, which showcases the 
effectiveness in sequence modeling tasks (glucose value prediction). 

Moreover, the Additive Attention module specializes in assigning adaptive weights to extract key components from high-dimensional features, an increase in prediction accuracy is reflected with the attention block (MAE/MAPE error drops by $12.1\%$/$10.0$\%). This quantitative result objectively showcases the effectiveness of our model architecture design.

As qualitatively shown in Figure \ref{fig:ablation_loss}, stacked LSTM blocks in the encoder will enhance representation learning in multi-scale compared with a single-block module, representing significantly lower loss values (e.g. orange/red lines). 
  In comparison with LSTM, 
  Bi-LSTM in the encoder captures features from both directions to learn more intricate temporal information about glucose levels (e.g. orange/blue lines).

\section{Conclusion}
This paper proposes a novel end-to-end glucose prediction model, TimeGlu, that utilizes only time series features to achieve state-of-the-art short-term glucose prediction. Four different baseline methods (DFS, Auto ARIMA, BATS, and TBATS) are also implemented in this paper for a thorough performance comparison. Through extensive experiments on the CGM Glucose and Colás dataset, TimeGlu achieves optimal glucose prediction accuracy without requiring additional personal data, demonstrating the potential of TimeGlu for future application in real-life diabetes management.

\bibliographystyle{IEEEtran}
\bibliography{bibtex/IEEEexample}

\end{document}